\documentclass{article}

\usepackage{arxiv}
\usepackage{graphicx}
\usepackage{amsmath}
\usepackage{amsfonts}
\usepackage{subfig}
\usepackage{xcolor}
\usepackage{xspace}
\usepackage{url}
\newcommand{\HV}{\ensuremath{\operatorname{HV}}\xspace}
\newcommand{\DP}{\ensuremath{\Delta_2}\xspace}
\newcommand{\w}{\ensuremath{\mathbf{w}}\xspace}
\newcommand{\A}{\ensuremath{\mathcal{A}}}
\newcommand{\Span}[1]{\ensuremath{\operatorname{span}\{#1\}}\xspace}

\newcommand{\BF}[1]{
	\relax
	\ifmmode
	\ifcat\noexpand#1\relax 
		\boldsymbol{#1}     
	\else
		\mathbf{#1}
	\fi
	\else
		\textbf{#1}
	\fi
}

\graphicspath{{figures}}

\begin{document}
\title{On Statistical Analysis of MOEAs with Multiple Performance Indicators}
%
%
\author{
\And
Hao Wang\\
LIACS, The Netherlands\\
h.wang@liacs.leidenuniv.nl
\And
Carlos Igncio Hernández Castellanos \\
Cinvestav-IPN, Mexico \\
chernandez@computacion.cs.cinvestav.mx
\And 
Tome Eftimov\\
Jo\v{z}ef Stefan Institute, Slovenia\\
tome.eftimov@ijs.si
}

\maketitle              
\begin{abstract}
Assessing the empirical performance of Multi-Objective Evolutionary Algorithms (MOEAs) is vital when we extensively test a set of MOEAs and aim to determine a proper ranking thereof. Multiple performance indicators, e.g., the generational distance and the hypervolume, are frequently applied when reporting the experimental data, where typically the data on each indicator is analyzed independently from other indicators. Such a treatment brings conceptual difficulties in aggregating the result on all performance indicators, and it might fail to discover significant differences among algorithms if the marginal distributions of the performance indicator overlap. Therefore, in this paper, we propose to conduct a multivariate $\mathcal{E}$-test on the joint empirical distribution of performance indicators to detect the potential difference in the data, followed by a post-hoc procedure that utilizes the linear discriminative analysis to determine the superiority between algorithms. This performance analysis's effectiveness is supported by an experimentation conducted on four algorithms, 16 problems, and 6 different numbers of objectives.

\end{abstract}

\section{Introduction}
A variety of Multi-Objective Evolutionary Algorithms (MOEAs) have been proposed to tackle the multi-objective optimization problem (MOP), e.g., NSGA-II~\cite{DebPratapAgarwalMeyarivan2002} and SMS-EMOA~\cite{BNE07}, which aim to optimize several objective functions simultaneously. In contrast to the single-objective optimization problems (SOP), for which the optimal solution is a single point in the search space, in MOPs, we seek the optimal trade-off between several objective functions, which is to approximate the Pareto front. Therefore, the difficulty of comparing MOEAs lies in quantifying approximation sets' quality to the Pareto front. Many performance indicators have been proposed and widely applied in many benchmarking and competition scenarios to determine a ranking among many tested algorithms, e.g., the hypervolume indicator~\cite{zitzler1999multiobjective} and the inverse generational distance (IGD)~\cite{coello2005solving}. Typically, in the performance analysis, we consider a set of performance indicators and report the empirical performance of MOEAs on each indicator separately. This treatment, however, makes it difficult to draw an overall conclusion on which algorithm is generally the best, since each performance indicator emphasizes on different aspects of the empirical performance, e.g., GD measures the distance of the approximation set to the Pareto front while the hypervolume indicator also considers the spread of approximation sets. Naturally, the performance indicators induce a multi-criteria decision-making problem which would pick the algorithm showing the best trade-off between performance indicators based on some decision criteria. 

The performance analysis becomes more complicated when it comes to the statistical significance: we usually apply a univariate non-parametric test (e.g., the Mann–Whitney U test) on each performance indicators independently for generating a ranking of algorithms, and aggregate such ranks over performance indicators when we would like to obtain an overview of the results~\cite{eftimov2017comparing}. This procedure would be sub-optimal and potentially fail to detect the significant difference between algorithms when the joint empirical distribution of performance indicators is multi-modal (each mode belongs to one algorithm) and, at the same time, its marginal distributions for each algorithm are overlapping. Therefore, we argue for the necessity of applying the multivariate testing procedure to the joint empirical distribution of performance data. This paper shall justify this proposal on the experimental data obtained on the combination of four algorithms, 16 problems, and six different choices for objective dimensionality (see Sec.~\ref{subsec:exp-setup} for the experimental setup).

This paper is organized as follows. In the next section, we will provide the background and related approaches for assessing the empirical performance of MOEAs. In Sec.~\ref{sec:method}, the proposed performance analysis procedure is discussed in detail. In Sec.~\ref{sec:results}, we demonstrate the outcome of our approach using the experimental data, followed by concluding remarks and future directions.

\section{Background}
    Here, we consider continuous multi-objective optimization problems of the form $\min_{x\in Q}F(x)$, 
      where $Q \subseteq \mathbb{R}^n$ and $F$ is defined as the vector of the objective functions $F:Q\to\mathbb{R}^k, F(x) = (f_1(x),\ldots,f_k(x))^\top$, and where each objective $f_i:Q\to\mathbb{R}$ is continuous. In this paper, we assume the Pareto order/dominance in the objective space $\mathbb{R}^k$ and denote the set of (Pareto) efficient points and the corresponding Pareto front by $P_Q$ and $F[P_Q]$, respectively.
    When $k > 4$, it is called a many-objective evolutionary problem. 

    \subsection{Evolutionary multi-objective optimization}
    Multi-objective evolutionary algorithms~\cite{Coello:07} are among the most widely-applied algorithms for solving MOPs:
    1) \emph{Dominance-based methods} guide the search points using the dominance relationship directly.
    In this class, one of the most popular methods is the non-dominated sorting genetic algorithm II/III (NSGA-II/III)~\cite{DebPratapAgarwalMeyarivan2002,DebJ14}, which is underpinned by the non-dominated sorting procedure for ranking the current solutions with respect to the Pareto dominance, and by measuring the crowding distance for diversifying the solutions.
    In this work, we will also test the SPEA2+SDE (Shift-Based Density Estimation for Pareto-Based Algorithms) algorithm which has been found to have good results in many-objective optimization problems~\cite{6516892}.
    2) \emph{Decomposition-based methods}~\cite{moubayed:14,ZL07} are built upon scalarization techniques, which can generate a good distribution of the Pareto front given a set of well-distributed scalarization problems.
    The most prominent algorithm in this class is MOEA/D (MOEA based on decomposition)~\cite{ZL07} that decomposes the original MOP into multiple different SOPs, each of which is tackled with the information from the neighboring SOP. The decomposition is usually realized via scalarizations, such as, the weighted sum, weighted Tchebycheff, or Penalty-based boundary intersection. 
    3) \emph{Indicator-based methods}~\cite{schuetze:16b} rely on performance indicators to assign the contribution to individual points.
    SMS-EMOA~\cite{BNE07} has received the most attention in this class. 
    
 

    \subsection{Performance assessment for multi-objective optimization}
    To assess the performance of multi-objective optimization, which is not a single solution but rather an approximation set, it can be analyzed with different criteria related to convergence and diversity. As we have previously mentioned, to do this, performance indicators are usually employed. These indicators are mathematical functions that map an approximation set to a real number~\cite{riquelme2015performance}. The most commonly used performance indicators are the hypervolume~\cite{zitzler1999multiobjective}, the generational distance~\cite{coello2005solving}, inverse generational distance \cite{coello2005solving}, epsilon~\cite{knowles2006tutorial}, spread \cite{deb2016multi}, and generalized spread~\cite{deb2016multi}. The selection of the performance indicator is a user preference and allows the user to select the performance indicator for which the newly developed algorithm performs the best, so in most cases such kind of comparisons can be biased. The influence of the selection of different performance indicators in comparisons studied has been investigated in~\cite{korovsec2020multi}.

    In some cases, one could be interested in using several performance indicators to assess the performance or to guide search of a MOEA. This could be helpful to improve the overall performance of an algorithm and help adapt it to different preferences from the user. For instance, in~\cite{10.1007/978-3-030-58115-2_14}, the authors proposed an ensemble approach which learns the best weights for a combination of five performance indicators. Further, advanced usage of the hypervolume indicators~\cite{7226795} over the space of indicators for automatic design of MOEAs.
    
    Recently, a Deep Statistical Comparison (DSC) approach has been presented for performing more robust statistical analysis for single-objective optimization that avoids the influence of outliers and small differences that exist in the data values~\cite{eftimov2017novel}. The main benefit of this approach is its ranking scheme, which is based on comparing distributions and not using only one descriptive statistic (either mean or median) to perform the analysis. The benefit of using the DSC approach for analyzing single performance indicator data has been shown in~\cite{eftimov2017deep}. Further, different ensembles have been proposed that combine the DSC results obtained for a set of performance indicators: an average ensemble~\cite{eftimov2017comparing}, a hierarchical majority vote~\cite{eftimov2017comparing}, and data-driven ensemble~\cite{eftimov2018data}. The average ensemble performs average of the DSC rankings obtained for different performance indicators for each algorithm on a specific problem. The hierarchical majority vote ensemble checks which algorithm wins (i.e. with regard to the DSC ranking) in the most performance indicators on each benchmark problem separately. The data-driven ensemble uses the preference of each performance indicator, which is estimated by its entropy. The PROMETHEE method \cite{brans2005promethee} is then used to determine the rankings.

\section{Methodology} \label{sec:method}
We typically consider multiple performance indicators at the same time to evaluate the performance of multi-objective optimization algorithms. When it comes to applying the statistical procedure to compare them, commonly, univariate hypothesis testing procedures are applied to each performance indicator, and the results are either presented separately or aggregated using simple methods. The potential risk of such approaches is that if the performance indicators are correlated and hence employing univariate tests on each indicator might fail to report significant differences among algorithms, which could only be discovered when we consider the joint distribution of performance indicators (see Fig.~\ref{fig:marginals} and its explanation in Sec.~\ref{subsec:multi-indicator} for an example). This consideration leads to our proposal that is to apply multivariate tests on the joint distribution of performance data.\footnote{The data files and the source code of this study can be accessed here: \url{https://github.com/wangronin/EMO21}.}

    
    
    \subsection{Our approach}
All ranking schemes of combining different performance indicators first compared the algorithms for each performance indicator separately and then combined the results from different comparisons using different heuristics. To avoid this, we extended the DSC ranking scheme, where the information obtained from several performance indicators is not analyzed separately but their joint distribution is compared. So instead of comparing one-dimensional data which is in the case of the DSC, we compared the high-dimensional joint distribution of the performance indicators. For this reason, we used the \textit{multivariate $\mathcal{E}$-test}~\cite{szekely2004testing}, which may be one of the most powerful tests available for high-dimensional data.

Further, we briefly reintroduce the DSC ranking scheme. Let us assume that $m$ algorithms are involved in the comparison. Let $\alpha$ be the significance level used for the \textit{multivariate $\mathcal{E}$-test} statistical test, where all $m\cdot (m-1)/2$ pairwise comparisons between the algorithms are performed and the p-values are organized in a $m \times m$ matrix, $N_i$, where $i$ is the benchmark problem: 
\begin{equation}
N_i[l_1,l_2]=\begin{cases}
p_\mathrm{value}, & l_1 \neq l_2\\
1, & l_1=l_2
\end{cases}
,
\label{eq:firsM}
\end{equation}
where $l_1$ and $l_2$ are different algorithms and $l_1,l_2 = 1,\dots, m$. 

Since multiple pairwise comparisons are made, the family-wise error (FWER) is controled by applying the \textit{Bonferroni correction}. Looking at the matrix $N_i$, we can say that is is always reflexive, and symmetric, but its transitivity should be checked. To do this, the matrix $N_i^{'}$ is introduced using the following equation:
\begin{equation}
N_i^{'}[l_1,l_2]=\begin{cases}
1, & N_i[l_1,l_2] \geq \alpha_\mathrm{X}/C_{m}^2\\
0, & N_i[l_1,l_2] < \alpha_\mathrm{X}/C_{m}^2
\end{cases}
.
\label{eq:secondM}
\end{equation}
Having the matrix $N_i^{'}$, if transitivity is satisfied, we can split the algorithms into disjoint sets of algorithms such that the algorithms that belong to the same set have the same joint distribution of the quality indicators and should be ranked as the same.

Formally, given a set of $m$ algorithms $\A = \{A_1, A_2, \ldots, A_m\}$ and a set of performance indicators $\mathcal{I} = \{I_1, I_2, \ldots, I_p\}$ (all subject to maximization, w.l.o.g.), this test returns a family of disjoint sets of algorithms that are significantly different from each other with respect to $\mathcal{I}$, namely $\mathcal{G} = \{G_1, G_2, \ldots\}$, where $\forall G \neq G' \in \mathcal{G}, G, G' \subseteq \A, G \cap G' = \emptyset$, and $\cup_{G\in \mathcal{G}} G = \A$. Based on this testing result, we proceed to determine which group is superior using the widely-applied \emph{Fisher's linear discriminative analysis} (LDA), which aims to find the best linear scalarization of performance indicators that maximizes the separation of the algorithm groups when scalarizing the points in each group. The rationale behind this approach is: having known that two groups are significantly different, we seek a subspace (representing the linear scalarization) onto which the projection of grouped performance points would respect this statistical significance to the maximal degree, such that in this subspace the decision of superiority for algorithms, even via simple descriptive statistics, e.g., the projected group mean, would not be affected by the randomness with high probability.
We shall denote by $\mathbf{w}$ the linear weights and use $\{\boldsymbol{\mu}_i\}_{i=1}^{|\mathcal{G}|}$ and $\boldsymbol{\mu}$ for the group means and overall mean of the performance values, respectively. The LDA procedure is to maximize the following separation measure: $S = \mathbf{w}^\top\boldsymbol{\Sigma}\mathbf{w} / \mathbf{w}^\top\mathbf{C}\mathbf{w}$, subject to $\mathbf{w} \in \mathbb{R}_{\geq 0}^p$, where $\boldsymbol{\Sigma} = \sum_{i=1}^{|\mathcal{G}|}(\boldsymbol{\mu}_i - \boldsymbol{\mu})(\boldsymbol{\mu}_i - \boldsymbol{\mu})^\top / |\mathcal{G}|$ is the sample covariance of the group means and $\mathbf{C}$ is the sample covariance of all performance points. 

Note that, it is necessary to enforce all components of $\mathbf{w}$ to be non-negative since we will loose the interpretability of the scalarization if performance indicators (subject to maximization) were combined using weights having different signs. Also, when this procedure fails to deliver a feasible $\mathbf{w}$ in analyzing our experimental data, we simply take the equal weights for all performance indicators. Having obtained the weights $\mathbf{w}$ from LDA, we could calculate the scalarized performance values, i.e., $\sum_{k=1}^p w_i I_k$, which will be referred as \emph{LD values} in this paper. To determine a ranking of algorithm groups, we propose to take the mean of scalarized points for each group and simply sort the resulting mean value.


 

    \subsection{Experimental setup} \label{subsec:exp-setup}
    We elaborate the experimental setting as follows. Algorithms: NSGA-II, NSGA-III, MOEA/D, and SPEA-SDE; Problems: DTLZ 1-7 and WFG 1-9, which are denoted as F1-F7 and F8-F16 in this paper, respectively; Number of objectives: 3, 5, 7, 10, and 15; Performance indicators: \DP and the hypervolume \HV; Function evaluation budget: 125,000; Independent runs: 20; We analyze the final population for each algorithm. The experiment is implemented using \textbf{Platemo}~\cite{TianCZJ17}. 

    \begin{figure}[!htpb]
        \centering
        \includegraphics[width=1\textwidth]{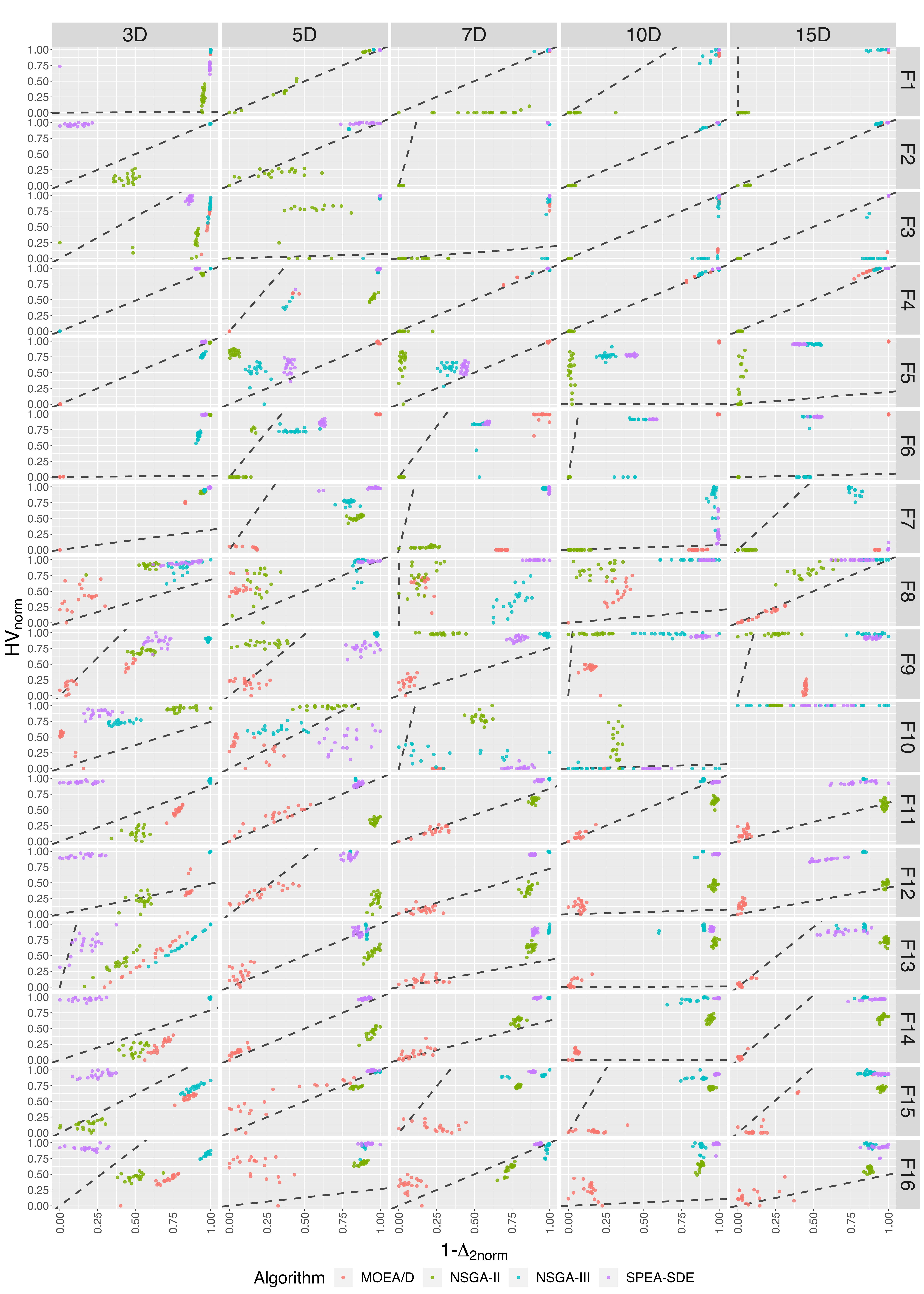}
        \caption{The normalized samples of two performance indicators, the hypervolume (\HV) and \DP, are shown for each pair of objective dimension and test function for four MOO algorithms, where weights \w of the best linear discriminative function is indicated by the dash line. The best discriminative function is determined by maximizing the separation between algorithm groups that are statistically significant (see Table~\ref{tab:epsilon-test-result}), after projecting them onto \Span{\w}.}
        \label{fig:scatter}
    \end{figure}

\section{Performance Analysis} \label{sec:results}
Firstly, the empirical performance of the tested algorithms are plotted in Fig.~\ref{fig:scatter}, where the performance values from $20$ repetitions are scatted in the performance space spanned by \HV and \DP. We re-scale the \HV and \DP values to the unit interval for each pair of objective dimension and problem, and additionally transform the normalized \DP value to $1-\DP$, such that the resulting values are subject to maximization. From the figure, it is obvious to see that variability of both performance indicators can change drastically across different algorithms, dimensions, and problems, for instance, both \HV and \DP are very stable for each algorithm on problem F2 across all dimensions while \HV exhibits a relatively huge dispersion on problems F8, F9, and F10 on dimensions $7$, $10$, and $15$.

    \subsection{Deep Statistical Comparison}\label{subsec:classical}
\begin{figure}[!htbp]
        \centering
        \includegraphics[width=1\textwidth]{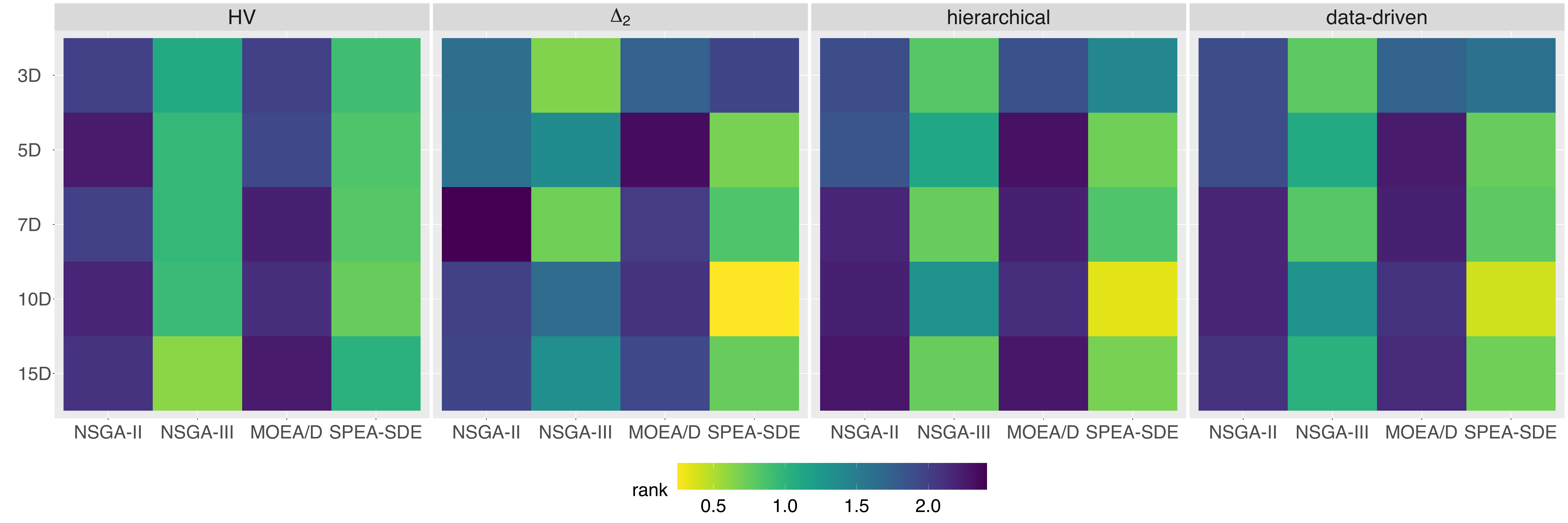}
        \caption{Averaged deep statistical comparison rankings of four algorithms over $16$ test problems with four different ranking schemes applied (from left to right): the DSC ranking on \HV, the DSC ranking on \DP, the DSC hierarchical ranking, and the data-driven DSC ranking.}
        \label{fig:mean-rank}
    \end{figure}

    Here, we applied the state-of-the-art statistical ranking approaches from the deep statistical comparison methodology\footnote{For applying the DSC ranking scheme, we used the \textbf{DSCTool}, which is developed as RESTful web services to the DSC ranking functionalities~\cite{EFTIMOV2020105977}.}, where the well-known two-sample Anderson-Darling (AD)~\cite{d1986goodness} test is taken to generate the ranks. In details, four statistical procedures are conducted: 1) the standard DSC ranking scheme on the \HV samples, 2) the standard DSC ranking scheme on the \DP samples, 3) the so-called DSC hierarchical ranking on both \HV and \DP, which determines the superiority of an algorithm by the majority vote from a set of performance indicators, and 4) a data-driven DSC ranking scheme on both \HV and \DP, in which we compute the entropy carried by each performance indicator and impose an entropy-based weight on rankings from each performance indicator when aggregating them. 
    We demonstrate the resultant DSC rankings as heatmaps in Fig.~\ref{fig:mean-rank}, where ranks obtained on each problem and objective dimension are averaged over 16 problems for a high-level comparison. When comparing the rankings from \HV to that from \DP, we observed some disagreements of those two indicators in deciding the ordering of algorithm, e.g., judging from both \HV, NSGA-III performs about the same as SPEA-SDE except that it shows better performance on $15$D, while SPEA-SDE takes the lead when the dimensionality gets higher in terms of \DP. Such disagreements are intrinsic to the performance indicators since \HV prefers solutions around the knee of the Pareto front while \DP prefers solutions close to the approximation set, which is chosen to be well-distributed along the Pareto front. Also, the hierarchical and data-driven DSC ranking schemes aggregate the rankings from each performance indicator to yield an overview of the empirical performance. Those two DSC ranking approaches exhibit, on our experimental data, quite consistent results across all dimensions of the objective space. It is worth noting that we always employ a univariate test to determine the rankings, even if aggregations are applied in those four comparison approaches.

    \subsection{Multi-indicator analysis}\label{subsec:multi-indicator}
    As proposed previously, we firstly applied the multivariate $\mathcal{E}$-test to compare the algorithms in a pairwise manner (with Bonferroni correction at a significance level of 5\%), which takes the 2D sample points in the performance space spanned by \HV and \DP and results in groups of algorithms that are significantly different from each other. We show, in Table~\ref{tab:epsilon-test-result}, the cases where there is at least one non-singleton set in the grouping. From the table, it is seen than algorithms NSGA-III and MOEA/D are often grouped together for smaller numbers of objectives while NSGA-III is statistically indifferent from SPEA-SDE when the objective dimension goes larger. 
    
\begin{figure*}[!htbp]
    \centering
    \subfloat[Performance space and marginals]{\includegraphics[width=.5\textwidth,trim=3mm 0mm 0mm 0mm,clip]{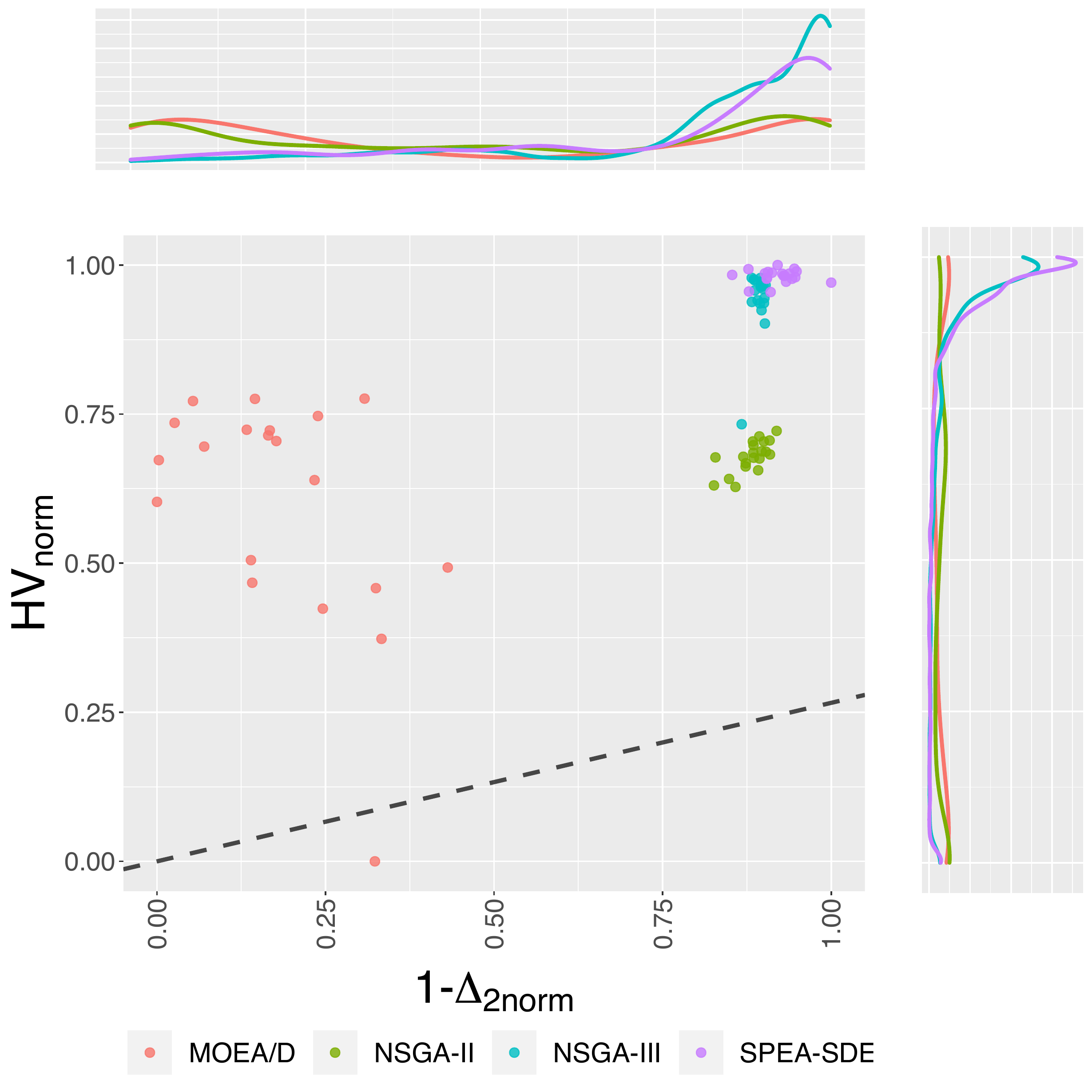}\label{fig:marginals}}
    \subfloat[Distribution of LD values]{\includegraphics[width=.5\textwidth,trim=10mm 10mm 0mm 10mm,clip]{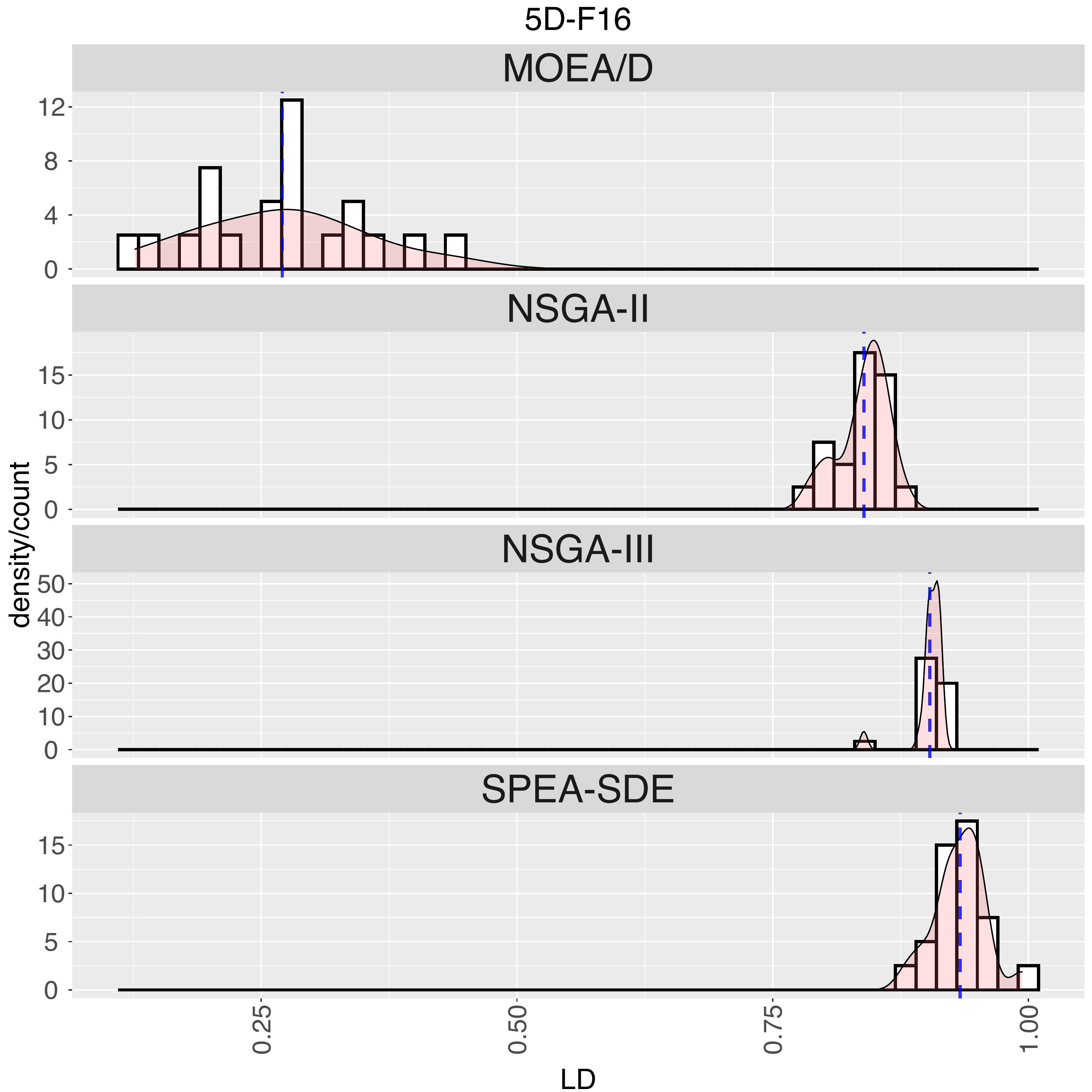}\label{fig:ld-values}}
    \caption{On the five-dimensional F16 problem, the performance points are shown in the scatter plot on the left, where the linear discriminative weights are indicated by the dashed line. Accompanying the scatter plot, the marginal distributions of \HV are largely overlapping for all algorithms (the same for \DP), in contrast to the clear difference of the LD values across algorithms (on the right).\label{fig:example-F16-5D}}
    \end{figure*}

    Taking the grouping information, we then proceed to apply the LDA procedure to obtain the optimal linear discriminative weights $\mathbf{w}$. In Fig.~\ref{fig:scatter}, the resulting weights are depicted as dashed lines for each pair of objectives dimensions and test problem, which varies largely from case to case. In addition, we zoom into the scenario of the five-dimensional F16 problem (Fig.~\ref{fig:marginals}) and also render the distributions of the LD values for each algorithm (Fig.~\ref{fig:ld-values}. Note that all groups are singletons in this case). The effectiveness of the proposed methodology lies in the fact that a univariate test applied on the marginal distributions of performance indicators would not lead to a significant discovery because the marginal distributions are largely overlapping for all algorithms (see the top and right sub-plots in Fig.~\ref{fig:marginals}, where the marginals are estimated using the kernel density estimation), as opposed to the outcome of the multivariate $\mathcal{E}$-test that all algorithms are significant different in this two-dimensional performance space. The LDA procedure, which could be considered as a post-hoc analysis to the multivariate $\mathcal{E}$-test, visualizes this statistical difference on the 1D subspace $\Span{\mathbf{w}}$ (the dashed line in Fig.~\ref{fig:marginals}): the 2D performance values, after projected onto $\Span{\mathbf{w}}$, are more distinguishable than the two marginals of performance values (see the histogram/density in Fig.~\ref{fig:ld-values}). Finally, to decide the ranking of algorithm groups, we could take the mean LD values since we have already gained the confidence on the group differences from the multivariate $\mathcal{E}$-test.
    \begin{table}[!htbp]
    \centering
    \setlength{\tabcolsep}{7pt}
    \begin{tabular}{l|l|c}
      \hline
       $k$ & $F(x)$ & $\mathcal{G}$ \\ 
      \hline
      3D & F1 & \{\{NSGA-II\}, \{NSGA-III, MOEA/D\}, \{SPEA-SDE\}\} \\ 
      3D & F3 & \{\{NSGA-II\}, \{NSGA-III, MOEA/D\}, \{SPEA-SDE\}\} \\ 
      3D & F4 & \{\{NSGA-II\}, \{NSGA-III, MOEA/D\}, \{SPEA-SDE\}\} \\ 
      3D & F8 & \{\{NSGA-II\}, \{NSGA-III, SPEA-SDE\}, \{MOEA/D\}\} \\ 
      5D & F1 & \{\{NSGA-II\}, \{NSGA-III, MOEA/D\}, \{SPEA-SDE\}\} \\ 
      5D & F2 & \{\{NSGA-II\}, \{NSGA-III, MOEA/D\}, \{SPEA-SDE\}\} \\ 
      5D & F3 & \{\{NSGA-II\}, \{NSGA-III, MOEA/D\}, \{SPEA-SDE\}\} \\ 
      5D & F4 & \{\{NSGA-II\}, \{NSGA-III, MOEA/D\}, \{SPEA-SDE\}\} \\ 
      7D & F3 & \{\{NSGA-II\}, \{NSGA-III, MOEA/D\}, \{SPEA-SDE\}\} \\ 
      7D & F8 & \{\{NSGA-II, MOEA/D\}, \{NSGA-III\}, \{SPEA-SDE\}\} \\ 
      10D & F2 & \{\{NSGA-II\}, \{NSGA-III, MOEA/D\}, \{SPEA-SDE\}\} \\ 
      10D & F8 & \{\{NSGA-II\}, \{NSGA-III, SPEA-SDE\}, \{MOEA/D\}\} \\ 
      15D & F3 & \{\{NSGA-II\}, \{NSGA-III, MOEA/D\}, \{SPEA-SDE\}\} \\ 
      15D & F10 & \{\{NSGA-II\}, \{NSGA-III, SPEA-SDE\}, \{MOEA/D\}\} \\ 
       \hline
    \end{tabular}
    \caption{The result of applying the multivariate $\mathcal{E}$-test on the bi-variate sample points $(\HV, \DP)$ with 5\% statistical significance and the Bonferroni correction. For each objective dimension and test problem, the groups of algorithms are significantly different from each other and we only show the cases where there is at least one non-singleton group in $\mathcal{G}$. \label{tab:epsilon-test-result}}
    \end{table}

\section{Conclusions and Future Work} \label{sec:conclusion}
In this paper, we delve into the question on how to utilize multiple performance indicators together in a statistical testing procedure to compare multi-objective optimization (MOO) algorithms, and how to best visualize and rank the performance data from multiple performance indicators when observing a significant difference between algorithms. For this purpose, we have chosen four evolutionary MOO algorithms, NSGA-II, NSGA-II, MOEA/D, and SPEA-SDE, which are independently executed for 20 times on DTLZ 1-7 and WFG 1-9 problems with the number of objectives $k\in \{3,5,7,10,15\}$. Subsequently, we compare those algorithms using the state-of-the-art Deep Statistical Comparison (DSC) method. Although it is easy to understand the resulting rankings, those are still computed using univariate testing procedures, which might fail to discover significant difference among algorithm, if the performance indicators exhibit a correlation structure. This comes from the fact that the selection of the performance indicators in DSC is still done by the user. Thus, we propose to directly apply a multivariate test ($\mathcal{E}$-test in this paper) in the multi-dimensional performance space spanned by those indicators. The result shows that this test could identify significant difference which would not be detected if we only apply univariate tests on the marginal distribution. Furthermore, we propose to visualize and rank the multivariate performance data by the linear discriminative analysis (LDA) method, which serves as a post-hoc procedure to the multivariate $\mathcal{E}$-test and seeks an one-dimensional subspace that maximizes the separation of algorithms when projected to this subspace. Having obtained statistical significance between algorithms, we could rank the algorithms in this subspace using simple descriptive statistics, e.g., the projected mean value.

For future works, it is important to evaluate the alternative multivariate testing procedures, e.g., especially nearest neighbor tests. Under the same significance level, we would, of course, choose the test with more statistical power. This question could be answered by conducting a power simulation on some experimental data. Also, we would like to study alternative post-hoc methods to LDA used here for determining the ranking of algorithms.
Finally, we would like to explore and quantify the loss of information when transforming high-dimensional data (i.e. approximation set) to one-dimensional data (i.e. quality indicator data) because this can also influence on the statistical comparison.

\section*{Acknowledgment}
Our work was financially supported by the Slovenian Research Agency (research core funding No. P2-0098 and project No. Z2-1867). 
We also acknowledge support by COST Action CA15140 ``Improving Applicability of Nature-Inspired Optimisation by Joining Theory and Practice (ImAppNIO)''. 

%
%
%
\bibliographystyle{unsrt}
\bibliography{mybibliography}
\end{document}